\documentclass[letterpaper, 10 pt, conference]{ieeeconf}
\usepackage{graphicx} 
\usepackage{booktabs}
\usepackage{multirow}

\usepackage{cite}

\usepackage{amsmath}
\usepackage{amssymb}
\usepackage{dirtytalk} 
\usepackage[table,xcdraw]{xcolor} 

\definecolor{purple}{HTML}{32006e}

\IEEEoverridecommandlockouts
\begin{document}
\title{Invariant Extended Kalman Filtering with Partial Orientation Measurement Integration: Theoretical Derivations and Application to Autonomous Surface Vessels}

\author{Derek R. Benham, Easton R. Potokar, Joshua G. Mangelson
  \thanks{This work was partially funded under Department of Navy awards N00014-24-1-2301 and N0014-24-1-2260 issued by the Office of Naval Research. Part of this work was also funded by the Simmons Research Endowment via the BYU IDR program.}
  \thanks{D.~Benham and J.~Mangelson are at Brigham Young University. E. ~Potokar is at Carnegie Mellon University. They can be reached via: \texttt{\{laser14, mangelson\}@byu.edu, potokar@cmu.edu}.
  }
}

\maketitle
\begin{abstract}
Autonomous surface vessels (ASVs) are increasingly vital for marine science, offering robust platforms for underwater mapping and inspection. Accurate state estimation, particularly of vehicle pose, is paramount for precise seafloor mapping, as even small surface deviations can have significant consequences when sensing the seafloor below. To address this challenge, we propose an Invariant Extended Kalman Filter (InEKF) framework designed to integrate partial orientation measurements. While conventional estimation often relies on relative position measurements to fixed landmarks, open ocean ASVs primarily observe a receding horizon. We leverage forward-facing monocular cameras to estimate roll and pitch with respect to this horizon, which provides yaw-ambiguous partial orientation information. To effectively utilize these measurements within the InEKF, we introduce a novel framework for incorporating such partial orientation data. This approach contrasts with traditional InEKF implementations that assume full orientation measurements and is particularly relevant for vehicles operating in a \say{semi-planar} environment, where the attitude is characterized by a dominant yaw rotation with limited roll and pitch variations. This paper details the developed InEKF framework; its integration with horizon-based roll/pitch observations and dual-antenna GPS heading measurements for ASV state estimation; and provides a comparative analysis against the InEKF using full orientation and a Multiplicative EKF (MEKF). Our results demonstrate the efficacy and robustness of the proposed partial orientation measurements for accurate ASV state estimation in open ocean environments.
\end{abstract}


\section{Introduction}
\label{sec:intro}

Autonomous surface vessels (ASVs) are experiencing rapid growth in utilization across a diverse range of marine applications including extensive environmental monitoring and critical infrastructure inspection. A key application within marine science is the creation of high-resolution bathymetric maps, which provide detailed topological representations of the seafloor using sonar scans from the vehicle. 
The accuracy of bathymetric maps is critically dependent on the precise state estimation of the ASV's pose at the surface. Even seemingly minor deviations in surface positioning and orientation can accumulate over the sonar beam's propagation path, leading to significant inaccuracies and distortions in the resulting seafloor maps. Thus, achieving robust and high-fidelity state estimation for ASVs is paramount for generating reliable bathymetric data.
\begin{figure}[t]
    \centering
    \includegraphics[width=\columnwidth]{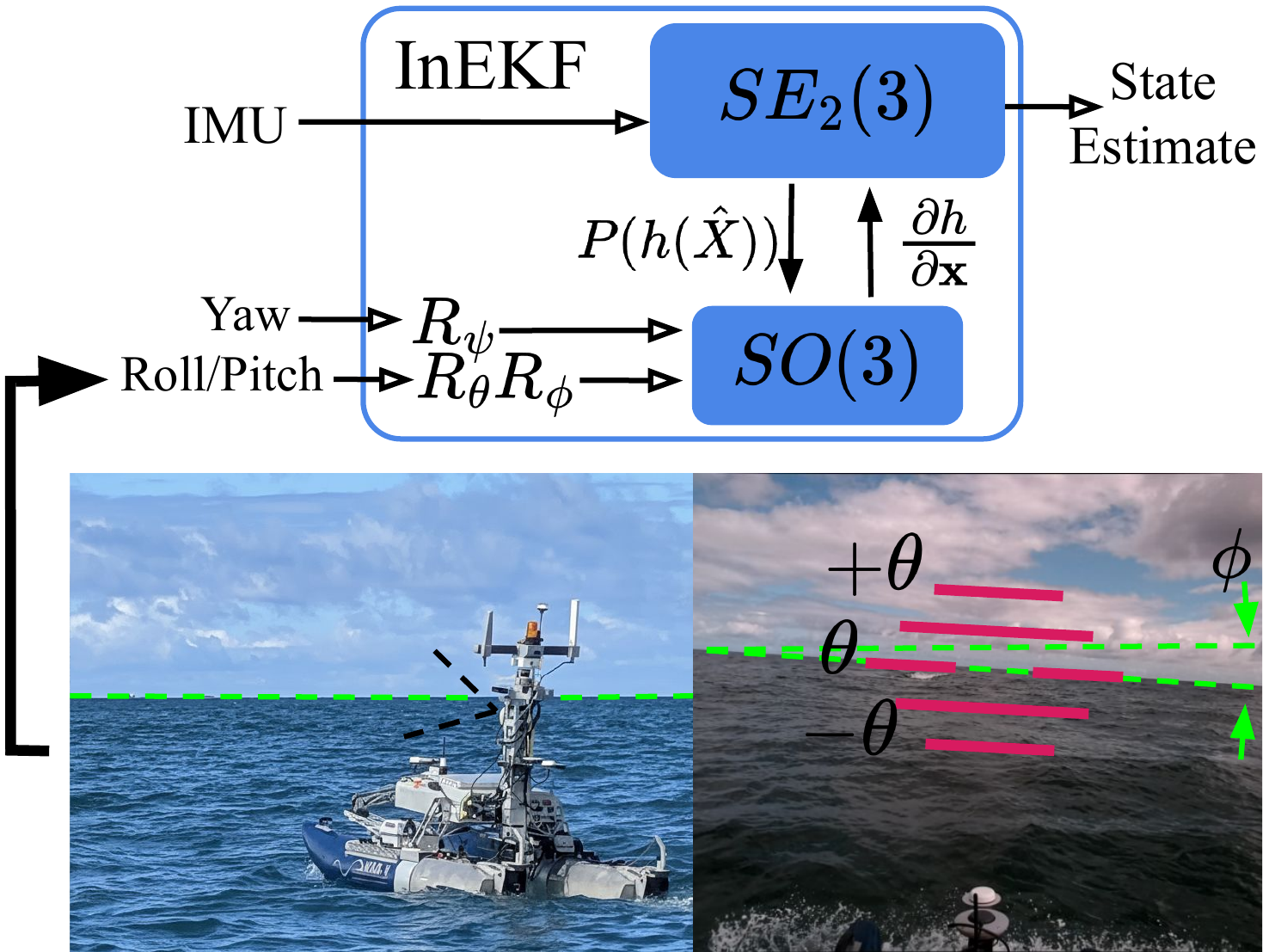}
    \caption{An overview of our novel proposed partial orientation measurement integration into the invariant EKF. The flow chart (top) shows how partial orientation measurements of roll ($\phi$) and pitch ($\theta$), or yaw ($\psi$) which reside in $SO(3)$ are bridged to the filter state of $SE_2(3)$ through the use of a planar frame and group homomorphisms. We then showcase the application of this framework to a simulated autonomous surface vessel (bottom left), where horizon observations from a forward-facing camera (bottom right) provide roll and pitch measurements for enhanced state estimation. 
    }
    \label{fig:highlevel}
\end{figure}

State estimation for ASVs in open ocean environments present inherent challenges. While ignoring roll and pitch or heave might suffice for some inland water applications where wave-induced motion is negligible, this simplification is not valid on the ocean and the full 3D orientation and position must be tracked. Given that the vehicle predominantly operates near a nominally level surface but experiences wave-induced roll and pitch, we introduce the concept of a \say{semi-planar} space. In this space, the attitude primarily involves rotation about the vertical $z$-axis, while the roll and pitch degrees of freedom are limited to moderate variations, a key constraint of our proposed methodology.
The absence of fixed features, a defining characteristic of the ocean, also presents a significant challenge for traditional landmark-based estimation approaches that rely on fixed landmarks for relative observations. Instead, the horizon serves as a consistently available visual reference and monocular cameras offer a practical and cost-effective solution for estimating the vessel's roll and pitch angles relative to it. Despite their robustness to certain environmental factors, these horizon-based measurements provide only partial orientation information, exhibiting limited sensitivity to yaw angle.

Invariant Extended Kalman Filters (InEKFs) offer a theoretically sound framework for state estimation on manifolds like the Special Orthogonal group $SO(3)$ for rotations and the Special Euclidean group $SE(3)$ for rigid body transformations. This approach overcomes many limitations found in traditional Extended Kalman Filters (EKFs). While current InEKF formulations require full orientation measurements, effectively integrating partial orientation measurements like horizon-derived roll and pitch, or heading from a dual-antenna GPS has yet to be developed.

To address this gap, this paper presents a novel framework for integrating partial orientation measurements within an InEKF. We propose a state estimation system for ASVs that fuses horizon measurements of roll and pitch with dual-antenna GPS heading by introducing an intermediary planar frame to simplify the estimation problem. We conduct a comparative analysis of the InEKF performance when using full orientation measurements versus our proposed decoupled approach with partial measurements, and further provide a performance comparison between our InEKF-based system and a traditional Multiplicative Extended Kalman Filter (MEKF) for ASV pose estimation.

The contributions of this paper include:
\begin{itemize}
    \item{A framework for integrating partial orientation measurements of \say{semi-planar} vehicles within an InEKF;}
    \item{A state estimation system for ASVs that fuses horizon measurements of roll and pitch with dual-antenna GPS heading, leveraging the proposed partial orientation measurement integration;}
    \item{A comparative analysis of the InEKF performance across simulated and real-world environments, evaluating full orientation measurements versus our proposed decoupled approach;}
    \item{A performance comparison between the proposed InEKF-based system and a traditional MEKF under both synthetic and operational conditions; and}
    \item{A quantitative evaluation of the accuracy of horizon-based attitude estimation, validated against industrial-grade ground truth in real-world maritime environments.}
\end{itemize}

The remainder of this paper is structured as follows. Section \ref{sec:related_work} reviews existing methodologies for ASV state estimation and the InEKF. Section \ref{sec:PO} introduces left and right invariant errors and the integration of partial orientation measurements such as roll and pitch, or yaw only within the InEKF. Section \ref{sec:horizon} presents the formulation for extracting roll and pitch from horizon observations using a monocular camera.  Section \ref{sec:results} details experimental validation and results, and finally, Section \ref{sec:conclusion} summarizes the key contributions and concludes the paper.

\section{Related Work}
\label{sec:related_work} 

\subsection{Autonomous Surface Vessel Localization Approaches}
Accurate state estimation is essential for ASVs to ensure reliable data collection. Planar models and filtering techniques, while suitable for calm inland waters, struggle to maintain accuracy when even small waves are present. \cite{hitz_hal-01174626} addressed those limitations by incorporating roll and pitch into their ASV state estimator. Their comparative study concluded that an EKF-based motion model—
following Fossen's propagation approach~\cite{fossen}
--combined with a complementary filter for attitude estimation produced the most stable velocity estimates. \cite{Hitz2016} further integrated their approach with LiDAR for coastal mapping, noting that while simple motion models improve velocity stability, they struggle with roll and pitch estimation.

Beyond simplified motion models, \cite{WANG2023115503} and \cite{Kjeldal_2018} explored more complex models for ASV state estimation, incorporating curvilinear motion and turn rate dynamics to improve robustness and stability.

In addition to traditional inertial navigation and GPS-based state estimation, LiDAR-based methods have been explored for ASV localization along coastlines and marinas. A comparative study by \cite{jmse11040840} analyzed various 3D LiDAR scan-matching techniques, while another study by \cite{berthing_estimation} investigated LiDAR-based state estimation for autonomous ships in constrained berthing scenarios. Consequently, these solutions are generally not applicable for open ocean environments where terrestrial references are absent.

\subsection{Horizon-Aided Attitude Estimation}
Several works have leveraged the horizon as a reference for attitude estimation in both aerial and marine domains. In aerial navigation, the horizon serves as a relative landmark for estimating roll and pitch. For example, \cite{6289875} fuses horizon-based measurements with gyro data to estimate drone attitude. Although other works achieve absolute roll and pitch estimation, they do so using georeferenced points \cite{5548331, 5507188}.

For marine vessels, the visual horizon offers a direct source of information without relying on georeferenced points
\cite{horizon_locking} introduced a lightweight semantic segmentation deep learning model 
for \say{horizon locking}, stabilizing images by transforming them to keep the horizon fixed relative to the camera despite vessel motion. In a similar approach, \cite{marine_vessel_attitude} employed a U-Net-based segmentation model to estimate the horizon line and extract roll and pitch. While their study concluded that horizon-based observations were more robust than raw IMU-based attitude estimation, it provided no rigorous numerical analysis of the resulting orientation errors. Furthermore, the performance was validated on a passenger ferry under minimal wave-induced motion, leaving the system's reliability in more dynamic or high-frequency sea states unverified.

\subsection{Invariant Kalman Filtering}
State estimation for systems evolving on nonlinear manifolds presents challenges for traditional Kalman filtering approaches, which rely on local linearizations that can introduce significant error. The InEKF addresses these issues by leveraging the Lie group symmetry of the system, allowing for a geometrically consistent representation of uncertainty. \cite{inproceedings_og} first introduced the idea of leveraging system symmetries within an EKF framework, leading to the development of the InEKF. Unlike the standard EKF, which linearizes around an arbitrary state, the InEKF ensures that estimation errors evolve on the Lie algebra, where process models can be shown to be linear for a specific class of systems~\cite{7523335}.

Lie groups provide a natural framework for robot state estimation, as they allow for a compact representation of both rotational and translational motion while preserving group structure. A comprehensive introduction to these mathematical structures can be found in \cite{sola2018micro}. Many estimation problems, including attitude estimation and robotic localization, can be formulated on matrix Lie groups such as $SE(2)$, $SE(3)$, and their higher-dimensional extensions. In this work, we specifically consider the matrix Lie group $SE_2(3)$, which extends $SE(3)$ to incorporate linear velocity as an explicit state variable, enabling a more structured treatment of IMU-based inertial navigation.

Subsequent work to \cite{inproceedings_og} has explored both theoretical extensions and practical applications of InEKF. \cite{5400372} derived both left- and right-invariant forms of the filter and applied them to a velocity-aided attitude estimation problem, demonstrating improved performance over traditional filtering methods. Introduced in \cite{hartley2020contact}, the notion of an \say{imperfect InEKF} provides a method to track and compensate for inherent internal biases in IMU sensors, offering improved performance over standard InEKFs even though these biases do not naturally reside within the same Lie group structure as the primary state. Recent advancements by \cite{barrau2022geometry} addressed this limitation, demonstrating accurate IMU bias tracking within the InEKF framework by reformulating it as a two-body problem.

The InEKF has since been applied across various robotics domains. \cite{hartley2020contact} extended the framework to contact-aided state estimation for legged robots. \cite{Potokar2021InvariantEK} applied the InEKF to underwater navigation and developed a method to incorporate single-axis position measurements while preserving the filter’s structure. By leveraging the Woodbury identity \cite{woodbury1950inverting}, they showed how to assign infinite covariance to unmeasured states (e.g., $p_x$ and $p_y$) while still allowing meaningful updates to a depth measurement ($p_z$). 
To make the InEKF more accessible, \cite{easton_tutorial} and \cite{mythesis} provide intuitive introductions, clarifying theoretical foundations and offering practical implementation insights.

\section{Partial Orientation Measurements within a ``Semi-Planar'' InEKF}
\label{sec:PO}
For a detailed derivation of the InEKF, we direct the reader to \cite{easton_tutorial}.
Within this section, we prioritize clarity and conciseness by outlining the essential InEKF equations and operators. This foundation will allow us to illustrate: 1) how a typical InEKF integrates full-state sensor data, 2) how group homomorphisms can be applied to integrate full orientation measurements, and 3) crucially, how our novel modification enables the use of partial orientation measurements, which is the core innovation of this paper. 
\subsection{Theoretical Primer for a Full-State Observation}
Invariant errors on a Lie group manifold exist in two forms: left ($\eta^l$) and right ($\eta^r$), defined as

\begin{equation}
\begin{aligned}
\eta^l \triangleq X^{-1}\hat{X} \\
\eta^r\triangleq\hat{X}X^{-1}\label{eq:lie_error_def},
\end{aligned}
\end{equation}
where $X, \hat{X}\in\mathcal{G}$ represent the true and estimated states on the Lie group $\mathcal{G}$. 
The group errors $\eta^l$ and $\eta^r$ can be mapped to the Lie algebra $(\mathfrak{g})$ using the logarithmic map $\mathcal{G}\rightarrow\mathfrak{g}$. The exponential map performs the inverse mapping $\mathfrak{g}\rightarrow\mathcal{G}$.
We define the vectorized left and right errors as defined in the Lie algebra as: 
\begin{equation}
\xi^l\triangleq \log(\eta^l)^\vee \label{eq:xi_def}
\end{equation}
\begin{equation}
\xi^r\triangleq \log(\eta^r)^\vee,
\end{equation}
where the vee operator ($^\vee$)  converts a skew-symmetric matrix in the Lie algebra to a vector, and the hat (or wedge) operator ($^\wedge$) performs the inverse transformation.

Solving for the state $X$ using the left error definition gives 

\begin{equation}
X=\hat{X}\exp(-\xi^{l\wedge}).
\end{equation}

In this paper, we implement a Left-InEKF on $SE_2(3)$, which exclusively tracks the left-invariant error $\eta^l$, defined in \eqref{eq:lie_error_def}.

Unlike a standard EKF, which linearizes the system around the current state estimate, the InEKF linearizes around the invariant error $\xi$, ensuring that estimation errors evolve predictably in the Lie algebra. Given a full-state measurement, the measurement model ($z$) and the predicted measurement ($\hat{z}$) can be defined as
\begin{equation}
\begin{aligned}
z&=X\exp(-\mathbf{w}^\wedge_m)\label{eq:measurement_model} \\
\hat{z}&=\hat{X},
\end{aligned}
\end{equation}
where $X, \hat{X}\in  SE_2(3)$ and the measurement noise $\mathbf{w}_m\sim \mathcal{N}(0,M)$ is modeled in the Lie algebra. The innovation is then defined as 
\begin{equation}
V^l=\log(z^{-1}\hat{z})^\vee. \label{vl_def_start}
\end{equation}
Substituting the measurement model for full state measurements \eqref{eq:measurement_model}, the innovation becomes
\begin{equation}
V^l=\log(\exp(\mathbf{w}^\wedge_m)X^{-1}\hat{X})^\vee. \label{v_l_step}
\end{equation}
With substitution from equations \eqref{eq:lie_error_def} and \eqref{eq:xi_def}, \eqref{v_l_step} simplifies to  
\begin{equation}
\begin{aligned}
V^l &= \log(\exp(\mathbf{w}^\wedge_m) \eta^l)^\vee \\
    &= \log(\exp(\mathbf{w}^\wedge_m) \exp(\xi^{l\wedge}))^\vee.
\end{aligned}
\end{equation}
Exponentials can be combined using the Baker-Campbell-Hausdorff first-order approximation \cite{Hall_2015}, leading to  
\begin{equation}
\begin{aligned}
V^l &\approx \log(\exp(\mathbf{w}^\wedge_m + \xi^{l\wedge}))^\vee \\
    &\approx \mathbf{w}_m + \xi^l. \label{vl_def_final}
\end{aligned}
\end{equation}
Linearizing $V^l$ about $\xi^l$ leads to $H=I$.

While this simplified model holds for full-state observations, in many real-world applications the full state cannot be observed and we are limited to orientation measurements, which may be complete or, more often, partial. 

\subsection{Orientation Measurement Integration}
Integrating orientation measurements into state estimation frameworks presents several unique considerations.
For instance, consider a scenario where a complete $SO(3)$ orientation measurement is given by $z=R\exp(-\mathbf{w}_{m})$, with its estimate $\hat{z}=\hat{R}$, despite the state being represented in $SE_2(3)$. To address this mismatch, group homomorphisms provide a systematic way to incorporate $SO(3)$ measurements into an InEKF formulated for $SE(3)$ or $SE_2(3)$ states.

A group homomorphism is a function $h: \mathcal{G} \rightarrow \mathcal{H}$ that respects the underlying group structure. Formally, for all $\mathcal{X,Y \in G}$, the following conditions must be satisfied:
\begin{equation}
\begin{aligned}
h(\mathcal{X} \circ \mathcal{Y}) &=h(\mathcal{X})\circ h(\mathcal{Y})\\
h(I_\mathcal{G})&=I_\mathcal{H} \\
h(\mathcal{X}^{-1})&=h(\mathcal{X})^{-1}.
\end{aligned} \label{group_h}
\end{equation}

These equations express the key properties of a group homomorphism: the preservation of the group operation, the identity element, and inverse elements, respectively.

In the case of the special orthogonal and euclidean groups, the $\circ$ operator is standard matrix multiplication. 
Based on these definitions, the group homomorphism $h:SE_2(3)\rightarrow SO(3)$ holds, and is simply $h(\mathcal{X})=R_\mathcal{X}$ where $R_\mathcal{X}$ is the rotation component of the state. In effect, the full-state measurement model, when projected through this homomorphism, becomes $z=h(Xexp(-\mathbf{w}_m))$, and the corresponding estimated measurement is $\hat{z}=h(\hat{X})$.
In this projected state, the same innovation can be applied as defined in \eqref{vl_def_start}, and the linearized measurement Jacobian $H$ becomes the Jacobian of the homomorphed measurement function $h$ with respect to the $SE_2(3)$ state. This Jacobian, denoted as $\mathbf{J}_h=\frac{\partial h}{\partial\mathbf{x}}$, will be a $3\times9$ matrix $\begin{bmatrix}I_{3\times3} && 0_{3\times6}\end{bmatrix}$. 
By projecting to $SO(3)$ via the homomorphism, we can effectively integrate full $SO(3)$ orientation measurements within our $SE_2(3)$ InEKF framework using the established innovation and Jacobian forms.

Challenges arise however when measurement provide only partial orientation information, such as roll and pitch alone or just yaw, rather than a complete orientation estimate. To overcome this issue, we propose two different approaches with examples for integrating a roll and pitch measurement, and yaw only.

\subsection{Partial Orientation Measurement Integration}
\subsubsection{Roll and Pitch Measurement}
Consider a sensor that provides absolute roll and pitch in world coordinates. In this instance, yaw is an unobservable parameter. We will begin by applying a group homomorphism ($h$), converting our state from $SE_2(3)$ to $SO(3)$. Following this, we define a new planar frame $R_p$ that consists of the X-then-Y set of rotations. 
Let $R_{p} \in SO(3)$ be a reference frame that is only rotated with respect to the world frame around the vertical $z$-axis. 
\begin{equation}
R_{p}^W = R_z(\psi) = \begin{bmatrix}\cos\psi && -\sin\psi && 0 \\ \sin\psi && \cos\psi && 0 \\ 0 && 0 && 1 \end{bmatrix} \label{eq:R_z}.
\end{equation}
The rotation between $R_p$ and the body frame can be formulated as a combination of roll ($\phi$) and pitch ($\theta$) rotations around the $x$ and $y$ axis respectively, 
\begin{equation}
R_{x}(\phi) =\begin{bmatrix}1 && 0 && 0 \\ 0 && \cos\phi && -\sin\phi \\ 0 && \sin\phi && \cos\phi \end{bmatrix}
\end{equation}
\begin{equation}    
R_{y}(\theta) =\begin{bmatrix}\cos\theta && 0 && \sin\theta \\ 0 && 1 && 0 \\ -\sin\theta && 0 && \cos\theta \end{bmatrix}.
\end{equation}
We will denote these roll and pitch rotations as $R_b^{p} = R_y(\theta)R_x(\phi)$. By left multiplying $R_{p}^W$ by $R^{p}_b$ we can construct the full rotation $R_b^W$. 

To formulate a roll and pitch sensor within the group structure we will define a function $\text{RollPitchProjection}(R)$ ( or $P(R)$ for short) that maps from the full set of rotations $SO(3)$ to the subset of X-then-Y rotations $R^p_b$.

This can be done by extracting the yaw value $\psi=\text{atan2\,}(R_{1,0}, R_{0,0})$ \cite{bernardes2022quaternion}, recreating the $R_z(\psi)$ matrix, and left multiplying it's transpose by the current belief state $\hat{R}^{W}_b$, giving us the robot's orientation in the planar frame $\hat{R}_b^{p}=P(\hat{R}_b^W)=R_z(\hat{\psi})^T \cdot \hat{R}_b^W$. Using this new function, the measurement model can now be defined as $z=P(h(Xexp(-\mathbf{w}_m)))$, where both the group homomorphism and $P(R)$ functions are applied. To represent the measurement and its estimate within $SO(3)$ in a simplified form, we can write $z=R_b^p$ and $\hat{z}=\hat{R^p_b}=P(\hat{R}^W_b)$.

The measurement $z$ and measurement belief $\hat{z}$ can now be treated as full orientation measurements for calculating the innovation. The linearization will remain the same as the full orientation measurement ($H=\begin{bmatrix} I_{3\times3} && 0_{3\times6} \end{bmatrix}$). The lack of information about yaw rotation in this measurement will be accounted for by effectively making the associated uncertainty infinitely large.  This approach to handling infinite uncertainty is discussed further in \ref{inf_uncertainty}.

While this approach provides a framework for incorporating partial pose measurements, it is important to acknowledge its limitations. 
Since rotation matrices do not commute, the elements of $\xi$ in the Lie algebra $\mathfrak{so}(3)$ act as generators of infinitesimal rotations, meaning that a sequence of small rotations about different axes are inherently coupled. When the rotation error is small, the components of $\xi=\begin{bmatrix} \xi_1 , \xi_2,\xi_3\end{bmatrix}^T$ approximately correspond to deviations in roll, pitch, and yaw respectively. In particular, small errors in roll ($\sim\xi_1$) and pitch ($\sim\xi_2$) introduce a small but non-zero yaw error ($\sim\xi_3$). due to the non-commutativity of rotations. This approximation holds when roll and pitch variations are small relative to a nominal planar frame—such as a vehicle operating near a level surface—where the influence of higher-order rotation effects remains negligible. 
Consequently, our \say{semi-planar} InEKF is well-suited for agents operating primarily near a nominally level surface while allowing for moderate roll and pitch variations.
While a more thorough analysis is needed to formally characterize this limitation, our system has demonstrated strong empirical performance in experiments with roll and pitch variations up to $\pm5$ degrees. Additional linearization testing indicates that it may remain effective for variations up to $\pm30$ degrees, though this remains to be fully validated. 

\subsubsection{Yaw Measurement}
An absolute heading ($\psi$) measurement provides rotation around the world $z$-axis. While this is similar to a magnetometer reading, the nature of the information obtained is fundamentally different. A magnetometer measurement typically provides a reference for the global $x$-axis in the body frame, inherently containing roll and pitch information, and easily integrates into the InEKF framework. In contrast, a pure yaw measurement provides no inherent information about roll and pitch angles and, unlike magnetometer readings, offers limited direct constraints on the full 3D orientation state.

If the agent operated in $SO(2)$, a yaw measurement could be represented similarly to a magnetometer, aligning with the world $x$-axis. However, because the vehicle operates in three-dimensional space, a map between $SO(3)\rightarrow SO(2)$ is not valid under the group homomorphism in \eqref{group_h}, as this would ignore roll and pitch dependencies inherent in the full $SO(3)$ representation.

We again define the planar frame as $R_p^W=R_z(\psi)$. 
As a dual-antenna GPS provides an absolute yaw measurement about the global $z$-axis, the measurement model in $SO(3)$ can be defined as $z=R^W_p\exp(-\mathbf{w}_\psi)$ and $\hat{z}=\hat{R}^W_p$. Here, $\hat{R}^W_p$ is created by extracting the estimated yaw value $\hat{\psi}$ from $\hat{R}$ and constructing an $SO(3)$ pose that represents only a rotation about the $z$-axis, see equation \eqref{eq:R_z}. This allows the measurement to again be incorporated as a full orientation measurement with the same innovation and linearization as before where $V^l=\log(z^{-1}\hat{z})^\vee$. 

While we have shown that a yaw-only measurement can be approximated as a full orientation measurement for filter updates, this simplification does not inherently capture the uncertainty inherent in this partial observation.

\subsubsection{Handling uncertainty}\label{inf_uncertainty}
Although partial orientation measurements can be incorporated into the filter as full orientation measurements, doing so introduces errors due to improper handling of uncertainty. The change of frame causes the yaw component of a roll and pitch measurement to appear nearly zero, while the roll and pitch components of a yaw measurement become exactly zero. As a result, the filter is incorrectly reinforced in its current belief for these states, increasing the risk of divergence.

To mitigate this issue, we assign a high uncertainty to the components that should be ignored, informing the filter that these measurements should be disregarded completely. Rather than using arbitrarily large covariance values, which can lead to numerical instability, we implement the analytical \say{infinite covariance} approach described in \cite{Potokar2021InvariantEK}.

In the InEKF measurement update, the Kalman gain $K$ is computed using $K=\hat{\Sigma}H^TS^{-1}$, where the innovation covariance is $S=H\hat{\Sigma}H^T+\hat{R}^TM\hat{R}$. Here, $\hat{R}$ represents the current orientation belief, and $M$ is the measurement covariance. Attempting to disregard a measurement by assigning an infinite value to an element in $M$ or by zeroing out the corresponding row in the measurement Jacobian $H$ can both result in a non-invertible innovation covariance $S$, preventing the calculation of $K$.

To address numerical challenges, an analytical solution for $S^{-1}$ is derived. To illustrate this, we consider the partial orientation measurement consisting of roll and pitch, with the measurement covariance matrix given by $M=\text{diag}(\sigma^2_\phi, \sigma^2_\theta, L)$, where $L\rightarrow\infty$ represents infinite variance in yaw, and the variances of roll and pitch are $\sigma_\phi^2$ and $\sigma_\theta^2$. 
By defining $\tilde{\Sigma}=H\hat{\Sigma}H^T$, the innovation covariances can be expressed as $S=\tilde{\Sigma}+\hat{R}^TM\hat{R}$, where $\hat{R}$ represents the current orientation estimate. To obtain an analytical form for $S^{-1}$, the Woodbury matrix identity \cite{woodbury1950inverting} is utilized that states that for any $n\times n$ matrices $A$ and $B$, $(A+B)^{-1}=A^{-1}-A^{-1}(AB^{-1}+I)^{-1}$. The following derivation applies this identity to formulate an expression for $S^{-1}$ that remains well-conditioned as $L$ approaches infinity.
\begin{equation}
\begin{aligned}
S^{-1}&= \lim_{L\rightarrow\infty} \Bigl(
\tilde{\Sigma}+\hat{R}^TM\hat{R}\Bigl)^{-1} \\
&=\lim_{L\rightarrow\infty}\tilde{\Sigma}^{-1}-\tilde{\Sigma}^{-1}\Bigl(\tilde{\Sigma}\Bigl(\hat{R}^T
\left[\begin{smallmatrix}
\sigma^2_\phi & 0&0\\0&\sigma^2_\theta&0\\0&0&L
\end{smallmatrix} \right]
\hat{R}\Bigl)^{-1}+I\Bigl)^{-1} \\
&=\tilde{\Sigma}^{-1}-\tilde{\Sigma}^{-1}\Bigl(\tilde{\Sigma}\Bigl( \hat{R}^T\lim_{L\rightarrow\infty}
\left[\begin{smallmatrix}
\frac{1}{\sigma^2_\phi} & 0&0\\0&\frac{1}{\sigma^2_\theta}&0\\0&0&\frac{1}{L}
\end{smallmatrix} \right]
\hat{R}\Bigr)+I\Bigr)^{-1} \\
&=\tilde{\Sigma}^{-1}-\tilde{\Sigma}^{-1}\Bigl(\tilde{\Sigma}\Bigl(\hat{R}^T
\left[\begin{smallmatrix}
\frac{1}{\sigma^2_\phi} & 0&0\\0&\frac{1}{\sigma^2_\theta}&0\\0&0&0
\end{smallmatrix} \right]
\hat{R}\Bigr)+I\Bigr)^{-1}
\end{aligned}
\end{equation}
This derivation enables the direct incorporation of partial orientation measurements into the estimation process under a closed form solution, making it suitable for real-world estimation tasks where only partial orientation data may be available.

\section{Horizon Observation as a Partial Orientation Measurement}
\label{sec:horizon}
For an ASV navigating the open ocean, the horizon provides a consistently available visual cue for orientation. In this section we define a framework for utilizing the horizon to observe roll and pitch information using a monocular camera.

\subsubsection{Horizon Detection}

We employ a lightweight computer vision pipeline for horizon extraction, though more advanced methods could be substituted. The process begins with image undistortion to ensure rectilinear line detection. Next, the Line Segment Detector algorithm developed by \cite{Grompone_lsd} detects prominent line segments within the undistorted image. Near-vertical segments, corresponding to features like clouds, wave crests or undistortion artifacts, are filtered out. The longest remaining segment is selected as the horizon observation—a choice that, based on empirical validation, suffices for our simulation results.

\subsubsection{Roll Angle}

The roll angle is computed directly from the detected horizon line's orientation in the image plane. Specifically, it is given by $\text{roll}=\text{atan2\,}(\Delta_y, \Delta_x)$, where $\Delta_y$ and $\Delta_x$ represent the vertical and horizontal pixel differences between the endpoints of the detected horizon segment. This effectively captures the in-plane rotation of the horizon relative to the image's horizontal axis.

\subsubsection{Pitch Angle}

\begin{figure}[t]
    \centering
    \includegraphics[width=0.8\columnwidth]{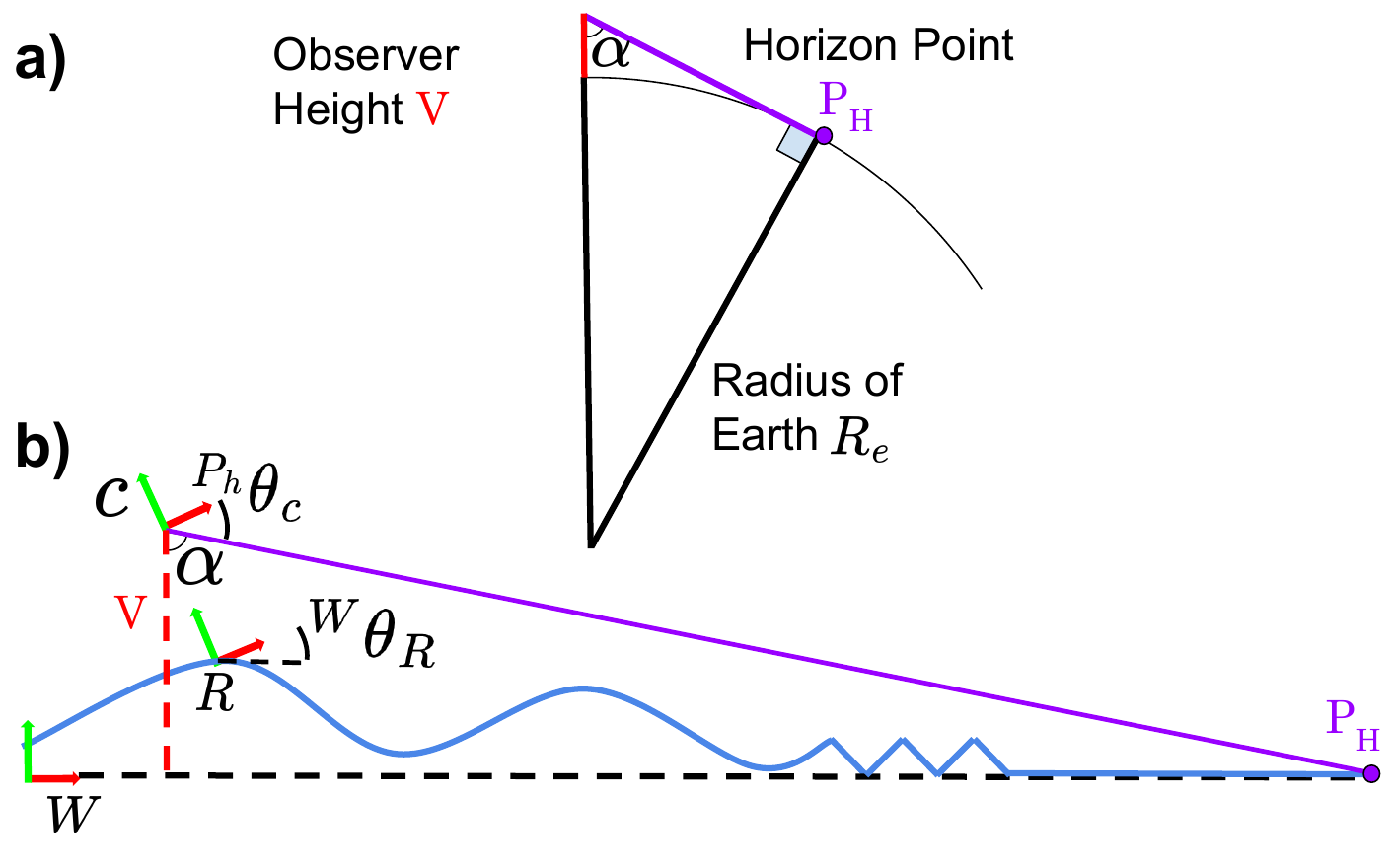}
    \caption{Labeled coordinate frames for estimating the pitch of an ASV by observing the horizon from a forward-facing camera. A) illustrates the geometric relationship used to calculate the inclination angle based on the observer's height above the Earth's surface and the Earth's radius. B) shows how the vehicle's pitch angle is then estimated using this calculated inclination angle and the relative declination of the observed horizon within the camera's field of view. 
}
    \label{fig:horizon_pitch}
\end{figure}

Estimating pitch ($\theta_R$) is geometrically more involved and is illustrated in Fig. \ref{fig:horizon_pitch}. The total angular depression of the horizon ($\alpha$) due to Earth's curvature, depends on the observer's height ($V$) above sea level and Earth's radius ($R_e$). This inclination $\alpha$ can be calculated as $\alpha=\arcsin(\frac{R_e}{R_e+V})$. The camera, however, observes a horizon inclination $\theta_c$ relative to its local horizontal. To derive the vessel's pitch angle ($\theta_R$) with respect to the true horizontal, we account for both the observed camera inclination and the horizon depression using $\theta_R=\alpha-\theta_c-\frac{\pi}{2}$. This equation compensates for both camera pitch and horizon depression, yielding an estimate of the vessel’s pitch relative to the geodetic horizontal.

\subsubsection{Camera Model and Declination Angle}

The declination angle $\psi$ relative to the camera plane is determined using the pinhole camera model. The horizon point, $p_{h_y}$, is defined as the vertical pixel coordinate where the horizon intersects the image centerline. To correct for skew, we adjust $p_{h_y}$ by subtracting $c_y$, the $y$-coordinate of the principal point in the camera calibration matrix. The angle $\psi$ is then computed using $\text{atan}2(p_{h_y}, f_y)$, where $f_y$ is the focal length in the y-direction.

The value of $V$ depends on the pitch and heave of the vehicle. However, we found that varying the camera height between one and three meters above the water surface results in less than a $0.02^\circ$ change in pitch, so we fix $V$ to be the camera height above the draft line.

\section{Results and Discussion}

\label{sec:results}
The results and discussion of our experiments are presented below. We start by detailing the simulation environment, then provide accuracy and convergence results of our filter and comparisons against an InEKF without our partial orientation measurements and the MEKF \cite{koch2020relative, mekf}.
Subsequently, we detail the real-world experimental setup off the coast of O'ahu, Hawaii, demonstrating how physical field data reinforces the simulation findings and validates the framework's robustness under practical maritime conditions.

\subsection{Simulation Environment}
To evaluate the performance of the proposed InEKF framework under controlled yet realistic marine dynamics, we utilized the high-fidelity HoloOcean maritime simulator \cite{holoocean}. This Unreal Engine-based environment permits the modeling of complex surface conditions, including FFT-based wave dynamics provided in recent updates to the platform \cite{romrell2025previewholoocean20}.

For these trials, we selected the WAM-V 16 ASV as the primary simulation platform. This choice ensures parity between our virtual and physical experiments; the WAM-V 16 is natively supported within HoloOcean and closely mirrors the dynamics of the WAM-V 8 platform used for our real-world data collection off the coast of O'ahu. This alignment allows for a direct comparison between simulated performance and field-observed behavior.

In the simulation environment, the WAM-V 16 model was configured to replicate the sensor suite used during our real-world data acquisition. This included: a forward-facing, calibrated monocular camera; an IMU to measure angular rates and linear accelerations; and a dual-antenna GPS receiver capable of measuring position and heading. To ease the jump between simulation to real-world experiments we employed a ROS2 wrapper of HoloOcean developed by \cite{meyers_ros2} along with the Python ROSbags library.

To establish a baseline dataset, the ASV was commanded to navigate a square pattern with 15-meter sides. This 3-minute mission provided a continuous stream of sensor measurements paired with simulator accurate ground truth states for filter validation.
Monte Carlo simulations were subsequently generated offline by adding independent, repeatable sensor noise to the recorded sensor data. The available sensors and noise parameters for this process are detailed in Table \ref{tab:sim_params}.

\begin{table}[]
\small\sf\centering
\caption{Simulation Parameters\label{tab:sim_params}}
\renewcommand{\arraystretch}{1.25}
\begin{tabular}{lrrll}
\toprule
Sensor        & Noise Std. Dev & Rate (Hz) \\
\midrule
Gyro        & $4.6\!\times\!10^{-4} ~\frac{\text{rad}}{s}~\frac{1}{Hz}$     & 200       \\
Accel        & $2.2\!\times\!10^{-3} ~\frac{m}{s^2}~\frac{1}{Hz}$        & 200       \\
Gyro Bias      & $4.4\!\times\!10^{-6}~\frac{\text{rad}}{s^2}~\frac{1}{\sqrt{Hz}}$  & 200       \\
Accel Bias     & $2.1\!\times\!10^{-4} ~\frac{m}{s^3}~\frac{1}{\sqrt{Hz}}$       & 200       \\
GPS XY Pos    & $1.4~ m$         & 1         \\
GPS Z Pos     & $2~ m$             & 1         \\
Heading       & $1^\circ$       & 1         \\
Roll \& Pitch & $0.15^\circ$     & 16        \\

\bottomrule
\end{tabular}
\end{table}

\subsection{Trajectory Comparison of State Estimators in Simulation}

To evaluate the performance of the InEKF with partial orientation measurements, its accuracy was compared against two alternatives: an InEKF utilizing a synthesized full orientation measurement and a standard MEKF. 
To construct the full orientation measurement required for a traditional InEKF update, the roll and pitch estimates derived from the vision pipeline were temporally aligned and fused with the 1~Hz global heading updates. Specifically, the roll and pitch measurement nearest in time to each heading update was selected to form a complete $SO(3)$ attitude. While this synthesis allows for a standard full-state update, it effectively downsamples the vision pipeline; the high-frequency roll and pitch information—available at 16~Hz—is discarded to match the 1~Hz heading rate. Consequently, this approach fails to leverage the majority of the available orientation data, resulting in significantly lower information throughput compared to the partial-measurement formulation. 
In contrast, the MEKF handles these asynchronous streams more naturally, as its additive error framework allows it to directly incorporate partial orientation measurements without the group-theoretic modifications required by the Invariant EKF.

\begin{figure*}[t]
    \centering
    \includegraphics[width=1\textwidth]{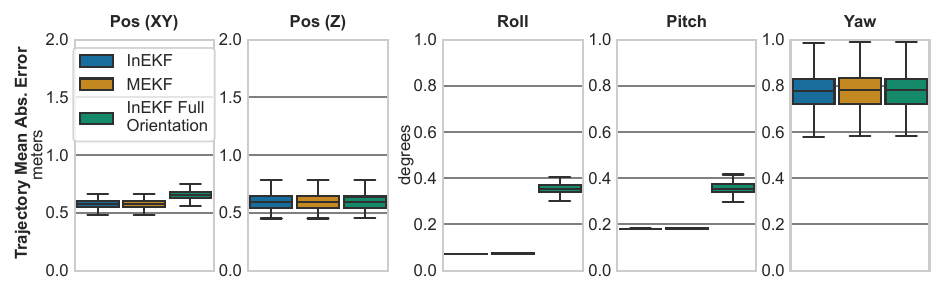}
    \caption{Average trajectory error comparison between our InEKF with partial orientation measurements and two benchmarks: a MEKF with the same measurements and an InEKF relying on lower-frequency full orientation measurements. The boxplots illustrate the distribution of errors across 100 Monte Carlo simulations. The effectiveness of our partial orientation measurement integration is evident; by fusing raw horizon observations directly as they arrive, our InEKF achieves comparable performance to the MEKF and notably outperforms the full-orientation variant. This performance gap highlights the advantage of the partial-measurement approach, which avoids the latency and reduced update rates required by methods that must first synthesize a complete orientation estimate before performing a filter update.}
    \label{fig:vanilla}
\end{figure*}

A series of 100 Monte Carlo simulations were conducted on the same trajectory, with each run initialized from randomized starting conditions. Each filter was applied to these trajectories, and absolute errors were recorded for each state, with the $L_2$ norm used for XY position errors. Fig. \ref{fig:vanilla} presents the average of the $L_2$ norm of the position error over the trajectory, as well as the average absolute error for the other states. 
While all three filters demonstrate comparable performance in height and heading estimation, the InEKF and MEKF, which utilize high-frequency roll and pitch measurements, exhibit superior roll and pitch as well as $xy$-position tracking. This improvement in position is a direct consequence of reduced attitude error; by more accurately resolving the gravity vector, these filters minimize the introduction of spurious horizontal accelerations into the state estimate.

These results demonstrate that the proposed approach effectively incorporates partial orientation measurements, achieving comparable accuracy to the MEKF in terms of final trajectory error, consistent with the findings of~\cite{hartley2020contact, Potokar2021InvariantEK}. This observation motivates a more detailed analysis of their convergence behavior in the following section.

\subsection{Filter Convergence Comparison in Simulation}
\begin{figure*}[t]
    \centering
    \includegraphics[width=1\textwidth]{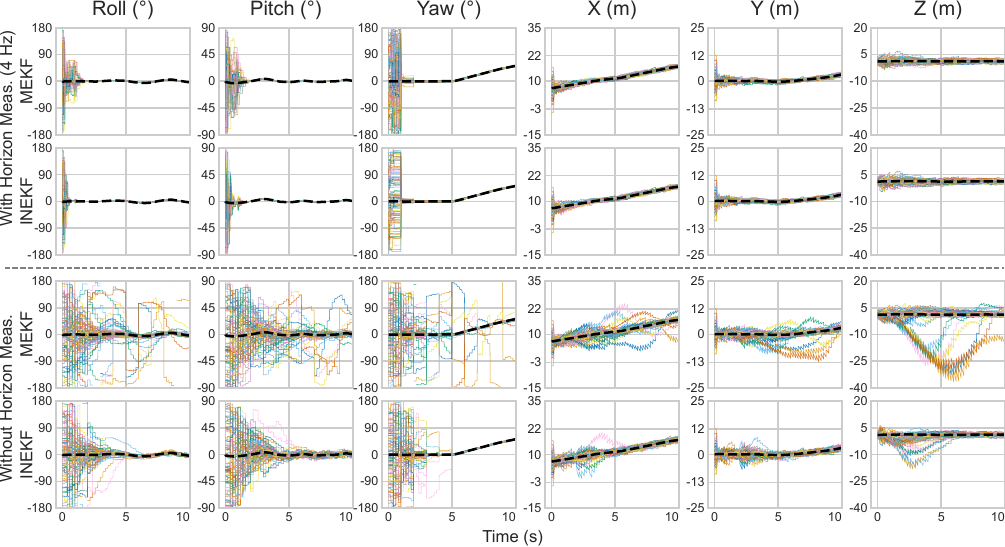}
    \caption{Convergence comparison of the InEKF and MEKF across six states (Roll, Pitch, Yaw, X, Y, Z) under two different measurement update strategies. Initial starting conditions for each trial were randomly sampled from a high-variance distribution (defined in Table \ref{tab:convergence_params}) to evaluate filter stability across a broad range of initialization errors. The top two rows illustrate the results with standard measurement updates including a 4 Hz horizon measurement, while the bottom two rows show the results when the horizon measurement is removed, relying solely on heading and GPS updates. Each subplot depicts 100 Monte Carlo simulations (light lines) and the ground truth (black dotted line) for 10 seconds of simulated data. These results demonstrate that our partial orientation measurement integration does not compromise the InEKF's superior convergence properties when compared to the MEKF.}
    \label{fig:convergence}
\end{figure*}

While the InEKF and MEKF may exhibit similar steady state accuracy, the speed and consistency with which an estimator converges to a reliable solution are paramount, particularly when faced with significant initial state errors. This is a known challenge for the MEKF, as its reliance on linearization around the current state error estimate means that poor initial guesses can lead to inaccurate linearizations and even divergence. 
In contrast, the InEKF defines its error in the Lie algebra through an invariant group action, decoupling the linearization from the current state estimate. This invariance allows the filter to correct large initial offsets and reliably return the estimate toward the true trajectory, even under highly uncertain initialization.
To specifically examine this behavior, we conducted 100 Monte Carlo simulations where initial state estimates were independently perturbed with Gaussian noise. The distribution parameters for these perturbations are specified in Table \ref{tab:convergence_params}, with the resulting convergence profiles illustrated in Fig. \ref{fig:convergence}.

In the first part of this experiment, we reduce the update rate of partial orientation roll and pitch measurements from 16 Hz (as in Fig. \ref{fig:vanilla}) to 4 Hz to better visualize the difference in the filters' convergence rate.
Despite the intentionally high initial error in the state belief, the presence of accurate and frequent roll and pitch measurements correct any potential linearization errors in the MEKF and both filters converge to the true solution. Notably, the InEKF exhibits a slightly faster convergence rate for the orientation states, though this improvement, while present, might be considered minor in the context of certain applications.

To make the distinction more evident, we conduct a second comparison where we remove roll and pitch measurements entirely, relying only on heading and GPS position updates. Despite the removal of horizon measurements, the InEKF still generally converged, albeit with reduced accuracy and an extended settling time. As anticipated due to its reliance on linearization about the state, the MEKF frequently diverged without the stabilizing effect of the frequent roll and pitch measurements. 

These results demonstrate that our integration of partial orientation measurements, such as roll and pitch, or yaw alone, preserves the InEKF's superior convergence properties over that of the MEKF. This provides a robust and reliable solution for pose estimation in autonomous systems where high-frequency full orientation measurements may not always be available. Ultimately, this experiment highlights the InEKF's resilience and potential for real-world deployment in scenarios with uncertain initial states.

\begin{table}[h]
\small\sf\centering
\caption{High Initial Noise Covariance\label{tab:convergence_params}}
\begin{tabular}{ll}
\toprule
State & Init Noise Std. Dev \\
\midrule
Orientation & $60^\circ$ \\
Velocity    & $4~ m/s$  \\
Position XY & $4~ m$    \\
Position Z  & $1~ m$    \\
\bottomrule
\end{tabular}
\end{table}

\subsection{Real-World Experiments}
While simulation provided a controlled environment, field trials are necessary to assess the system's performance under practical, potentially unmodeled maritime dynamics. 
To verify the mathematical validity of our partial orientation measurements integrated into the InEKF, and to evaluate the precision of using horizon-derived roll and pitch estimates, we deployed the vehicle off the coast of O'ahu to collect physical data for experimental validation.

In addition to benchmarking filter accuracy and convergence properties, we provide computational timing metrics to demonstrate the real-time viability of the system. 
Collectively, these results demonstrate that the proposed InEKF extension is a robust and computationally efficient solution for maritime state estimation, capable of bridging the performance gap between low-cost vision sensors and industrial-grade navigation systems.

\subsubsection{Data Collection and Trajectory}
Real-world data was collected off the coast of O'ahu, Hawaii, using a WAM-V 8 ASV. For evaluation, we partitioned a segment of the collected data that captures the ASV traveling away from the shore towards with an unobstructed view of the horizon. This specific segment covers a distance of approximately 575 meters over a duration of four and a half minutes.

\subsubsection{Instrumentation and Calibration}
Visual imagery was captured by a Luxonis OAK-D Long Range Camera. For the proposed method, only the global shutter RGB sensor was used in the state estimator.

Two IMUs were integrated on the vehicle: a low-cost BNO086 embedded within the OAK camera (OAK‑IMU) and a high‑grade SBG Systems Ellipse‑D (SBG‑IMU). The SBG Ellipse‑D includes a dual‑antenna RTK‑referenced GPS and provides an onboard Quaternion Extended Kalman Filter (QEKF) solution. We adopt the QEKF output, referred to as the SBG‑QEKF, as a high‑fidelity reference for comparison. While not considered absolute ground truth, all reported metrics describe deviations relative to this estimated trajectory. The intrinsic parameters of the IMUs were calculated using the Allan Variance method \cite{AllanVarianceRos}, and the extrinsics between the camera and IMU were jointly estimated using the Kalibr toolbox \cite{kalibr2,kalibr1}.

The RTK system reports an accuracy ($1\sigma$) of 1.4~cm in both $x$ and $y$, and a $0.35^\circ$ accuracy in yaw. Sensor data rates across the system are as follows: the SBG IMU operates at 200~Hz, the OAK IMU at 178~Hz, GPS position and heading at 1~Hz, and camera imagery at 16~FPS.

\subsection{Accuracy of Horizon-Based Roll and Pitch Measurements}

Prior to evaluating the InEKF using partial orientation measurements on real-world data, we first assess the validity of using real-world observations of the horizon to infer roll and pitch. Because data collection occurred exclusively on clear days with minimal horizon occlusion, the simple line segment detection algorithm previously employed in simulation was deemed sufficient. For more robust implementations of horizon detection, the reader is directed to the survey paper by ~\cite{horizon_survey}.

\begin{figure}[t]
    \centering
    \includegraphics[width=1\columnwidth]{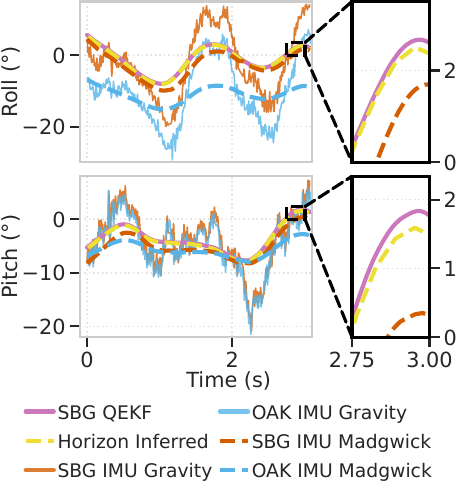}
    \caption{ 
    Estimated roll and pitch over a three-second interval for six methods: gravity-based estimates using the SBG-IMU and OAK-IMU, Madgwick-filtered estimates from both IMUs, raw horizon-based estimates, and the SBG-QEKF as a baseline. Gravity-based estimates degrade under external accelerations, and the Madgwick filter remains sensitive to bias—particularly in the OAK-IMU. Horizon-based estimates most closely align with the RTK-referenced, industrial grade IMU filter (SBG-QEKF), achieving sub-degree accuracy and demonstrating the effectiveness of using observations of the horizon to infer roll and pitch.
    }
    \label{fig:horizon_accuracy}
\end{figure}

To contextualize the accuracy of horizon-based estimates, we first compare them to static attitude estimates derived directly from the gravity vector. When the vehicle is stationary or moving slowly, the accelerometer primarily measures gravity, allowing roll and pitch to be inferred from the measured acceleration components using simple trigonometric relationships of roll, $\phi\!=\!\mathrm{atan2}(a_y, a_z)$, and pitch, $\theta\!=\!\mathrm{atan2}(a_x, g)$.
However, these gravity-based estimates degrade rapidly once external accelerations arise from vehicle motion.

As a second comparison, we consider the Madgwick filter~\cite{madgwick2010efficient}, which fuses accelerometer and gyroscope data to produce a smoothed estimate of the gravity direction. By integrating gyro rates while using accelerometer measurements to correct drift, the Madgwick filter provides a more stable attitude estimate than raw accelerometer measurements alone. However, it remains susceptible to gyro and accelerometer bias, and the uncertainty of its filtered output is not straightforward to incorporate into a higher-level estimator such as the InEKF.

The estimated attitudes over a representative three-second interval are shown in Fig.~\ref{fig:horizon_accuracy}. Among all methods evaluated, the horizon-based estimates exhibit the smallest deviation from the SBG-QEKF reference (an industrial grade IMU referenced by RTK GPS), achieving an RMS error of $0.18^\circ$ in roll and $0.21^\circ$ in pitch. A full comparison of each method relative to the SBG-QEKF baseline is summarized in Table~\ref{tab:horizon_eval}.

The motivation for using horizon observations in this context stems from their fundamental advantage over IMU‑based methods: they provide a direct geometric measurement of roll and pitch relative to the world frame, offering an absolute attitude reference that does not accumulate drift. In contrast, gravity‑based estimates break down in dynamic conditions due to external accelerations, and pre‑filtered approaches such as the Madgwick filter introduce additional model assumptions and uncertainty handling challenges when integrated into a higher‑level estimator.

The analysis further reveals that horizon-based pitch estimates typically exhibit higher errors than roll, a discrepancy rooted in how each state is geometrically extracted. Roll is derived from the slope of the detected horizon line; it is inherently robust to vertical offsets caused by atmospheric refraction, solar glare, or detection jitter, as the slope is invariant under vertical translation. In contrast, pitch is determined by the vertical position of the horizon within the image frame, rendering it highly sensitive to pixel-level disturbances. For the OAK-D LR camera, which features a vertical resolution of 480 pixels and a $56^\circ$ vertical field of view, a single-pixel shift results in a pitch fluctuation of approximately $0.12^\circ$.

\begin{table}[h]
\small\sf\centering
\caption{Roll and Pitch Estimator Accuracy Compared to SBG-QEKF\label{tab:horizon_eval}}
\begin{tabular}{lcc}
\toprule
\textbf{Method} & \textbf{Roll RMS} & \textbf{Pitch RMS} \\
\midrule
Observations of Horizon & 0.18$^\circ$ & 0.21$^\circ$ \\
SBG IMU Madgwick        & 1.78$^\circ$ & 1.51$^\circ$ \\
SBG IMU Gravity Vec     & 6.35$^\circ$ & 3.93$^\circ$ \\
OAK IMU Madgwick        & 9.96$^\circ$ & 2.4$^\circ$  \\
OAK IMU Gravity Vec     & 10.99$^\circ$ & 3.94$^\circ$ \\
\bottomrule
\end{tabular}
\end{table}

\subsection{Trajectory Comparison of State Estimators on Real Data}
\begin{figure*}[t]
    \centering
    \includegraphics[width=1\textwidth]{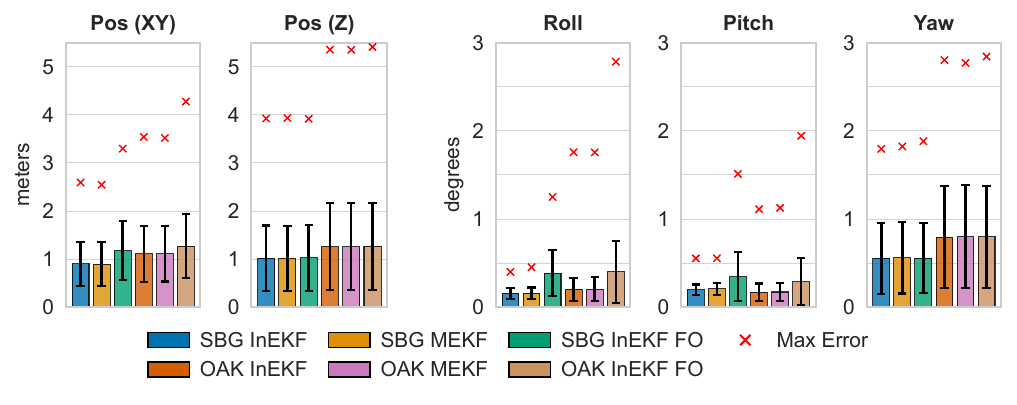}
    \caption{Average trajectory error comparison across high-grade (SBG) and low-cost (OAK) inertial hardware, evaluating our InEKF with partial orientation measurements against MEKF and full orientation InEKF benchmarks. The results illustrate the mean absolute state error across the 575-meter maritime trajectory, with error bars representing the distribution of residuals. These field results corroborate our simulated findings, showing that our partial orientation integration achieves comparable performance to the MEKF and significantly outperforms the full orientation variant. The latter is limited by the lower frequency of synthesized orientation solutions, whereas our method fuses raw horizon observations at the full camera framerate. Notably, the metrics demonstrate that horizon-based observations act as a hardware equalizer; while the OAK-based estimators exhibit higher variance due to sensor noise, their average roll and pitch accuracy remain competitive with the industrial SBG-based solutions.}
    \label{fig:rw_trajectory_error}
\end{figure*}

With horizon-based estimates now established as an accurate attitude measurement relative to high-fidelity benchmarks, the focus shifts to validating the comparative performance of the filter variants. We seek to demonstrate that the insights gained from simulation hold true under practical dynamics, specifically confirming that: 1) the MEKF and InEKF variants employing partial orientation measurements consistently outperform the traditional InEKF dependent on full orientation, and 2) the InEKF with partial orientation measurements retains its superior convergence properties relative to the MEKF, even in the presence of unmodeled environmental dynamics.

To evaluate these hypotheses across varying hardware capabilities, we compare the following estimators using measurements from both an industrial-grade SBG Ellipse-D and a low-cost BNO086 IMU, internally mounted inside the Luxonis OAK camera:
\begin{itemize}
    \item \textbf{SBG InEKF / SBG MEKF:} These filters integrate partial orientation measurements such as yaw only, or roll and pitch derived directly from the horizon observations.
    \item \textbf{SBG InEKF FO (Full Orientation):} A variant that relies on full orientation measurements by fusing the most recent horizon-based roll and pitch estimates with external heading into a single orientation measurement before processing.
    \item \textbf{OAK Variants (OAK InEKF, OAK MEKF, OAK InEKF FO):} Each of the aforementioned architectures is replicated using the noisier, low-cost measurements from the IMU in the OAK camera. These variants are used to assess the filter's resilience to hardware degradation and sensor noise.
\end{itemize}

All estimators are further constrained by GPS position and heading measurements with the noise parameters detailed in Table \ref{tab:sim_params}. This multi-tier comparison allows us to assess how closely the horizon-aided InEKF aligns with the commercial, RTK-referenced, SBG-QEKF solution across different price points of inertial hardware.

The experimental evaluation was conducted over the entire 575-meter trajectory, lasting approximately 4.5 minutes. All implemented filters were initialized using the position and heading estimated by the SBG-QEKF baseline, while velocities and roll/pitch were initialized to zero. To isolate steady state performance, the first 20 seconds of data are excluded from the error statistics. This window is intentionally conservative, as all filter variants typically achieve full convergence well within this timeframe (see \ref{sub:rw_convergence}).

The resulting performance, summarized by the Mean Absolute Error shown in Fig. \ref{fig:rw_trajectory_error}, corroborates the findings from our simulation studies. The MEKF and InEKF with partial orientation measurements demonstrate nearly identical performance across all states. Consistent with our previous observations, all filter variants show strong agreement in $z$-position and yaw estimation. However, the architectures that utilize partial orientation measurements once again exhibit superior accuracy in $xy$-position and orientation relative to the Full Orientation (FO) variant.

When comparing hardware performance, the SBG-based estimators predictably outperform the OAK-based implementations in overall position tracking, likely due to the superior accelerometer stability of the industrial unit. However, a notable parity exists in attitude estimation: the mean roll and pitch errors remain comparable across both hardware grades. This suggests that the horizon-tracking algorithm acts as a powerful equalizer; by providing a robust geometric constraint, the vision based observations effectively mitigate the higher noise floor of the consumer-grade IMU.

The primary distinction between the two platforms lies in their error distributions rather than their means. While both achieve similar average accuracy, the OAK-based filters exhibit significantly higher variance, with maximum roll and pitch errors reaching $1^{\circ}$ to $2^{\circ}$. This wider distribution reflects the decreased stability and stochastic noise inherent in consumer-grade inertial hardware, which the vision-based constraints can center, but not entirely suppress.

\subsection{Convergence Resilience Demonstrated with Real-World Data}
\label{sub:rw_convergence}

\begin{figure}[t]
    \centering
    \includegraphics[width=1\columnwidth]{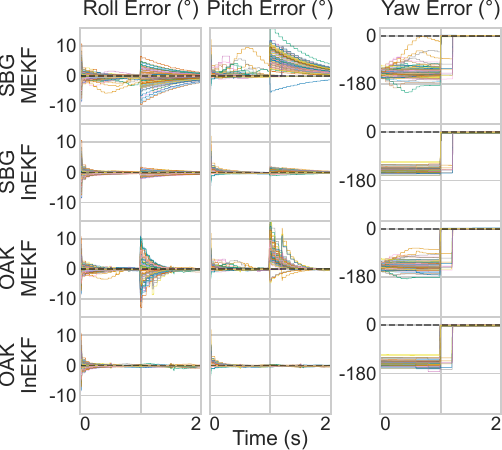}
    \caption{Average state error convergence for the InEKF and MEKF in roll, pitch, and yaw, aggregated across ninety 2-second segments for both SBG and OAK IMUs. To account for varying vessel states throughout the mission, each segment is plotted as the absolute error relative to the RTK-referenced filter provided by the industrial-grade IMU. All filters were initialized with an attitude at the identity, presenting a substantial initial yaw error around $150^\circ$. The results demonstrate that the InEKF converges significantly faster and with less oscillation than the MEKF across both hardware grades, validating its superior resilience to poor realistic initialization in real-world maritime conditions.}
    \label{fig:real_world_convergence}
\end{figure}

Despite the close agreement among the steady state trajectory estimates, the principal advantage of the InEKF over the MEKF remains its superior convergence behavior. 
To demonstrate this resilience in a practical real-world scenario, Fig. \ref{fig:real_world_convergence} illustrates the convergence trajectories for both filters, implemented on the SBG and OAK hardware.

To assess convergence, ninety 2-second segments were partitioned from the 4.5-minute trajectory. While the simulation included near-singular initializations (e.g., inverted vehicle poses), the real-world evaluation focused on practical \say{cold-start} scenarios. At each segment’s onset, the filters were initialized at the attitude identity (zero roll, pitch and yaw) and zero velocity, with position derived from the current GPS data.
This setup induced a substantial initialization error of approximately $150^\circ$ in yaw, alongside minor deviations of up to $10^\circ$ in roll and pitch. Despite this poor starting point, the comparison across these segments shows that the InEKF stabilizes rapidly, converging toward the true trajectory significantly faster than the MEKF for both the industrial and consumer-grade IMUs.

\begin{figure}[h]
    \centering
    \includegraphics[width=1\columnwidth]{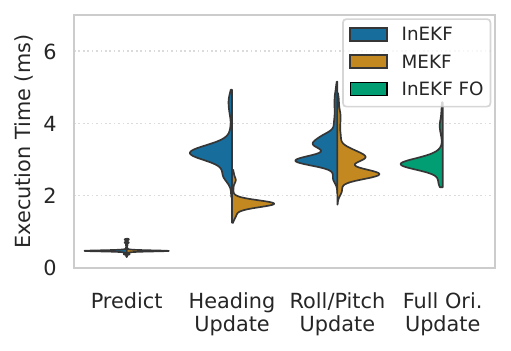}
    \caption{Computational execution times for the InEKF and MEKF variants across the 4.5-minute experimental trajectory. The violin plots illustrate the estimated probability density of processing latency for the inertial prediction ($\approx$ 200 Hz), horizon-based orientation update ($\approx$ 16 Hz), and global heading update ($\approx$ 1 Hz) functions. While a bimodal distribution is observed due to Python’s internal scheduling, all functions consistently execute in under 5 ms. This confirms that the framework maintains substantial computational headroom for real-time maritime navigation, even when implemented in an interpreted language.}
    \label{fig:timing}
\end{figure}

\subsection{Computational Performance Analysis}
To evaluate the operational feasibility of the proposed framework, we analyze the execution time of each estimator and the horizon-tracking vision pipeline. This analysis is critical for confirming that the state-independent convergence benefits of the Invariant EKF do not come at a prohibitive computational cost.

\subsubsection{Real-Time Execution and Filter Latency}
Execution times for the prediction and update functions were recorded over the entire 4.5-minute trajectory. Reflecting the multi-rate nature of the sensor suite, this analysis encompasses over 52,000 prediction steps driven by the 200 Hz IMU, 4,295 roll and pitch updates from the 16 Hz camera, and 260 yaw updates from the 1 Hz heading measurements. The violin plots in Fig.~\ref{fig:timing} illustrate the resulting probability distributions for each function call, providing a direct computational comparison between the InEKF and MEKF implementations. To ensure a deterministic evaluation of the core algorithmic complexity, all tests were performed on a consumer laptop with an Intel Core i7-8665U CPU (1.90 GHz, 4 Cores) with 16 GB of RAM. Linear algebra operations were strictly constrained to single-threaded execution by setting the \texttt{OMP\_NUM\_THREADS}, \texttt{MKL\_NUM\_THREADS}, 
and \texttt{OPENBLAS\_NUM\_THREADS} environment variables to 1. This configuration isolates the per-core computational cost of the state estimation logic from the overhead of multi-threaded library management.

Across the entire trajectory, the three sigma upper bound for any single function call remained under 5 ms. The bimodal distribution observed in the timing data is likely an artifact of Python's task scheduling and non-deterministic garbage collection. Although these results demonstrate real-time capability in a managed environment, a compiled C++ implementation would offer even greater latency reductions and more deterministic performance.

As expected, the InEKF update steps for partial orientation measurements exhibit slightly higher execution times compared to those of the MEKF. This is attributed to the increased complexity of the InEKF's measurement updates, which require transforming the attitude into an intermediary frame and employing the Woodbury matrix identity. Furthermore, the InEKF's measurement Jacobian ($H$) is consistently a matrix of size $3 \times 15$ for each update step, necessitating additional matrix multiplications. In contrast, the MEKF processes these as separate $1 \times 15$ and $2 \times 15$ Jacobians, which reduces the per-iteration computational cost.
The Full Orientation Update, shown on the right of Fig.~\ref{fig:timing}, is slightly faster than its partial measurement counterpart. This efficiency gain occurs because the Full Orientation variant bypasses the intermediary frame rotation and Woodbury identity, though it still utilizes the standard $3 \times 15$ Jacobian matrix. Despite these differences, all variations remain well within the real-time constraints of the system at their respective operational frequencies

\subsubsection{Vision Pipeline Throughput} 
The throughput of the integrated system is primarily governed by the computer vision horizon detection pipeline. Over 4000 callbacks of the image processing function yielded a mean execution time of 46~ms with a standard deviation of 7.2~ms. This equates to an average operational frequency of approximately 21.56 Hz, which is significantly higher than the 16 Hz camera framerate used during our field trials. These results indicate substantial computational headroom. Even with the slightly higher overhead inherent to the InEKF's group-theoretic updates and the use of an interpreted language, the system comfortably maintains real-time performance. Consequently, the framework is experimentally validated for deployment on autonomous surface vessels where low-latency state estimation is critical for safe navigation.

\section{Conclusion}
\label{sec:conclusion}
In this work, we derived a novel framework for integrating partial orientation measurements into the InEKF for \say{semi-planar} vehicles. We implemented this framework into a state estimation system that fuses horizon-derived roll and pitch measurements with dual-antenna GPS heading. Through repeated simulations, we demonstrate that our InEKF-based system utilizing high-frequency partial orientation measurements outperforms an InEKF relying on lower-frequency full orientation measurements. Furthermore, while the MEKF exhibited steady state accuracy comparable to that of our InEKF, under high initial state uncertainty the InEKF demonstrated superior convergence speed and stability to that of the MEKF. 

Real-world experiments conducted off the coast of O'ahu reinforced these findings. The horizon-derived attitude measurements achieved sub-degree accuracy ($0.18^\circ$ in roll and $0.21^\circ$ in pitch) relative to the industrial-grade RTK-referenced SBG-QEKF reference, effectively outperforming traditional IMU-based estimates. Notably, these vision-based constraints acted as a hardware equalizer; the partial orientation InEKF allowed consumer-grade sensors to achieve mean attitude accuracy competitive with industrial systems. Multiple trials initialized at the attitude identity further validated the InEKF’s superior convergence, showing rapid recovery from large state offsets where the MEKF exhibited prolonged settling times. Finally, timing metrics confirmed that both the Lie group updates and the vision pipeline are computationally efficient, maintaining substantial headroom for real-time deployment even in interpreted languages.

In the future, we aim to extend this state estimation framework to incorporate relative coastline observations, enabling robust and accurate deployment in complex littoral environments.

\bibliographystyle{IEEEtran}
\bibliography{ref}
\end{document}